\documentclass[
]{ceurart}

\usepackage{lmodern}
\usepackage{microtype}
\microtypesetup{expansion=false}
\usepackage{listings}
\usepackage{csquotes}
\usepackage{graphicx}  
\usepackage{subcaption} 
\usepackage{geometry}   
\begin{document}

\copyrightyear{2024}
\copyrightclause{Copyright for this paper by its authors.
  Use permitted under Creative Commons License Attribution 4.0
  International (CC BY 4.0).}

\conference{CHR 2024: Computational Humanities Research Conference, December 4–6, 2024, Aarhus, Denmark}


\title{Latent Structures of Intertextuality in French Fiction}

\subtitle{How literary recognition and subgenres are framing textuality}


\author[1,2]{Jean Barré}[
orcid=0000-0002-1579-0610,
email=jean.barre@ens.psl.eu,
url=https://crazyjeannot.github.io/]\cormark[1]

\address[1]{École normale supérieure - Université PSL, 45 rue d'Ulm, Paris, 75005, France}
\address[2]{LaTTiCe (Langues, Textes, Traitements informatiques, Cognition), 1 rue Maurice Arnoux, Montrouge, 92049, France}

\cortext[1]{Corresponding author.}

\begin{abstract}
Intertextuality is a key concept in literary theory that challenges traditional notions of text, signification or authorship. It views texts as part of a vast intertextual network that is constantly evolving and being reconfigured. This paper argues that the field of computational literary studies is the ideal place to conduct a study of intertextuality since we have now the ability to systematically compare texts with each others. Specifically, we present a work on a corpus of more than 12.000 French fictions from the 18th, 19th and early 20th century. We focus on evaluating the underlying roles of two literary notions, sub-genres and the literary canon in the framing of textuality. The article attempts to operationalize intertextuality using state-of-the-art contextual language models to encode novels and capture features that go beyond simple lexical or thematic approaches. Previous research \cite{Hughes_2012} supports the existence of a literary “style of a time", and our findings further reinforce this concept. Our findings also suggest that both subgenres and canonicity play a significant role in shaping textual similarities within French fiction. These discoveries point to the importance of considering genre and canon as dynamic forces that influence the evolution and intertextual connections of literary works within specific historical contexts.

\end{abstract}




\maketitle

\section{Introduction}
How has textuality been shaped over time? Can we account for the dynamics of influence and imitation in literary history? What roles have played the underlying structures, such as the literary canon or the literary genres? 

Intertextuality is the concept that centralizes these questions. It was introduced by \citet{kristeva_semeiotike_2017}, stating that “every text is a mosaic of quotations; every text is the absorption and transformation of another text." This perspective suggests that a text is no longer finite and that its complete interpretation and understanding must involve unraveling the network of textual relationships. A few years later, \citet{Barthes_1974} further develops the definition, stating that “every text is an intertext; other texts are present in it, at varying levels, in more or less recognizable forms: texts of the previous culture and those of the surrounding culture; every text is a fabric of past quotations." This metaphor of fabric, recalling that the etymology of the word “text" refers to the idea of weaving, illustrates that all literary creation is in reality a web of collective contributions. The text, conceived as a complex network of interwoven threads, represents a meeting point where various influences, voices, and cultural traditions interlace.

Intertextuality has been defined differently by successive researchers and can have multiple approaches\footnote{For an overview of the concept, see \citet{allen_intertextuality_2022}'s work.}. The main definition of intertextuality lies in the detection of explicit citation. Literary studies (computational or not) have naturally taken up tracking citations from one text to another to understand phenomena of influence or rewriting in literature (\cite{allen_plundering_2010}, \cite{B_chler_2012}, \cite{ganascia_2014}). A second approach to intertextuality, which we will call “weak" to contrast with the first, consists of a simple allusion or thematic and linguistic similarities between a given text and a set of other texts. Computational methods have enabled the development of this approach in recent years, with examples such as \citet{Ganascia_2020} who conducted automatic detection of textual fragments that evoke reuse due to their similarity by n-grams. This approach allows for the identification of intertextual relationships between texts that may not be immediately apparent through manual analysis. 

In this way, computational literary studies can be extremely useful for analyzing the intertextual network in large corpora of text. Researchers can now identify patterns and trends in the use of intertextuality across different genres, time periods, and cultural contexts. One example of this is the seminal paper by \citet{manjavacas_statistical_2020}, which proposes a thematic and lexical approach to intertextuality. By grouping authors and texts that share topical or lexical similarities, their aim was to evaluate whether it was possible to detect intertextual phenomena from a given source (i.e. the Bible) in a particular corpus. In contrast, our research views potential intertexts as endogenous to the corpus, and we will assess whether the most significant intertexts are also part of the literary canon or not. 


Before expanding on our experiment, we need to define two literary notions that may impact intertextuality: the first one is the literary canon. The dynamics of prestige, defined as an author or a book being included in school curricula or reviewed in a prestigious literary journal, can influence textuality. This is because the literary canon represents what is considered to be paradigmatic. Previous computational research has uncovered disparities in textual content between canonical and non-canonical works across various corpora and cultural backgrounds (\cite{algee-hewitt_canonarchive_2016}, \cite{underwood_distant_2019}, \cite{judith_brottrager_modeling_2022}, \cite{barre_operationalizing_2023}, \cite{barre_beyond_2023}). These studies have shown that formalist and transcendent definitions of the literary canon seem to be relevant across different genres and time periods. More interestingly, Underwood found an increase in the probability of a work being canonized over time, suggesting that canonical literary works constitute what \citet{altieri_idea_1983} terms a “cultural grammar." In other words, canonical works function as foundational texts shaping the norms, values, and conventions within a specific cultural tradition.

The second is the concept of literary genre. According to \citet{genette_theorie_1986}, one of the five dimensions of intertextuality (or transtextuality, as Genette refers to it) is architextuality\footnote{For an attempt to operationalize the concept at the passage level using a computational approach, see \citet{barre:hal-04559749}}, which is dedicated to literary genres. This dimension refers to the interconnectedness and interdependence of various texts within a literary genre, encompassing “all that sets the text in relationship, whether obvious or hidden, with other texts". Genette considers genre to be a crucial component of a work's overall meaning, arguing that the way in which texts reference and relate to one another can reveal important thematic and structural elements. \citet{jaus_toward_2010} supports this argument by introducing the concept of the “horizon of expectations" of the audience, which may lead authors to adhere to certain expected norms and styles. As a result, intertextuality is stronger between texts from the same genres, and genres constitute a structuring element for intertextuality.

These two concepts can be seen as structuring elements of intertextuality. Their impact on writing varies depending on their nature: unconsciously through the authority or influence of certain works, or consciously through literary subgenres. Our goal is to develop a methodology capable of automatically detecting the works and authors that have influenced the evolution of the intertextual network, and to determine which of our two concepts has the greatest impact. To achieve this, we rely on a massive corpus of over 12,000 French-language fiction.

As a proxy for prestigious literature, we will rely on novels that have been republished over time. We were assuming that books that have been republished multiple times have either sold well or are considered important enough to be reread or included in school curricula. The goal is to evaluate the role of canonical works in shaping the literary tradition. Our hypothesis is that the influence of canonical works on the intertextual network is more enduring than the pace of change in the archive, which is heavily represented by texts from popular literary sub-genres. In other words, we expect that canonical works will continue to shape the literary tradition over a longer period of time, while the popularity and influence of the others may be more short-lived.

\vspace{0.3cm}
\textbf{Outline of the paper}

The structure of the paper is as follows: We start with a detailed description of the method we used to model intertextuality (\autoref{modeling}), including the corpus description (\autoref{corpus}), the metadata construction (\autoref{metadata}) and the operationalization pipeline (\autoref{operationalization}). Then, we present the results in \autoref{results}, including an evaluation of the method (\autoref{evaluation}), and an analysis of the main results (\autoref{main_res}), the individual (\autoref{indiv_res}) and collective ones (\autoref{collective_res}). The article concludes with a discussion and the perspectives opened up by this research (\autoref{discussion}).

\section{Modeling Intertextuality}\label{modeling}

\subsection{Corpus}\label{corpus}

\begin{figure}[!ht]
    \centering
    \includegraphics[width=10cm]{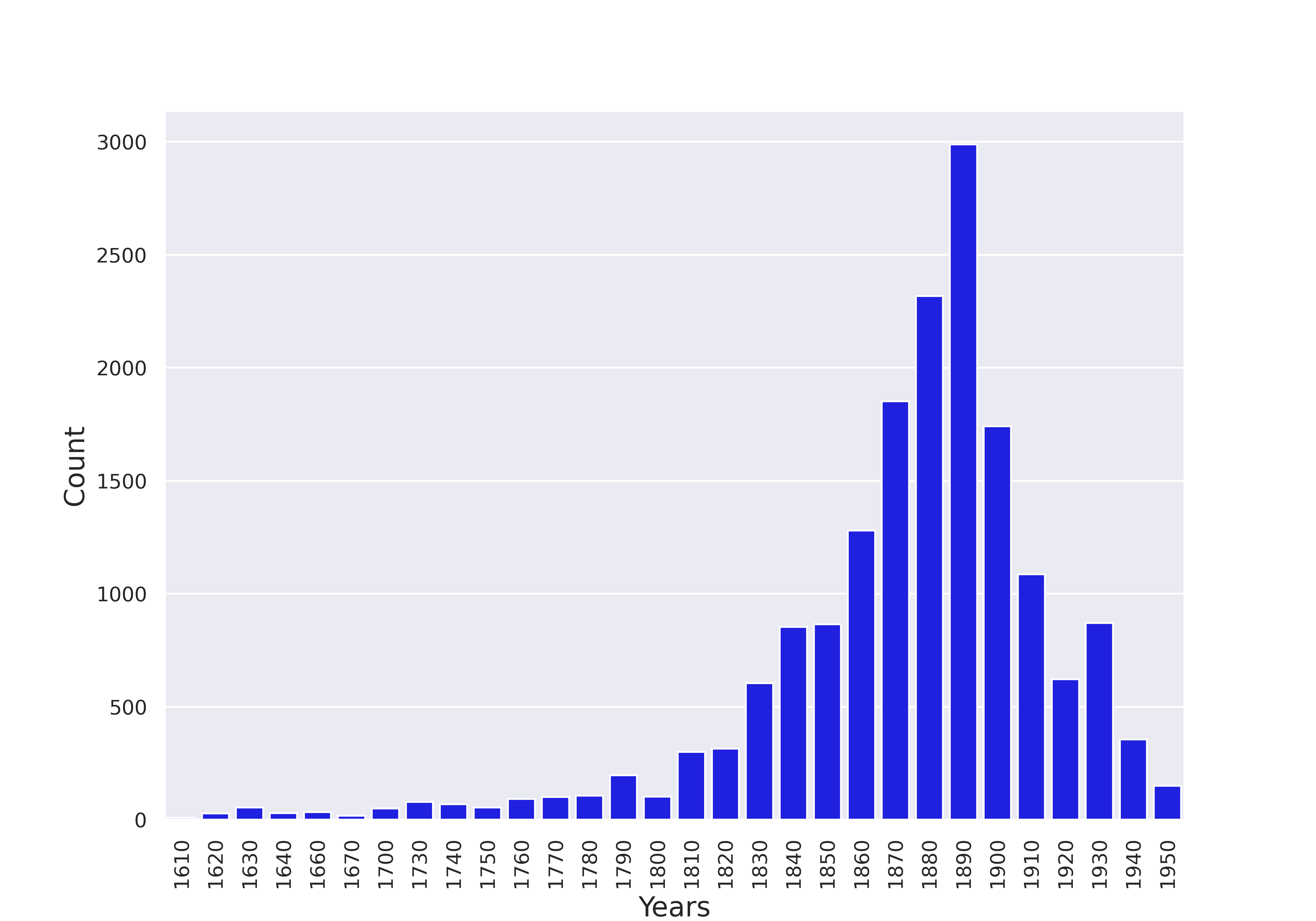}
    \caption{Number of novels over time}
    \label{fig:corpus_bar}
\end{figure}

Our corpus relies on a subset of the collection “Fictions littéraires de Gallica" \cite{langlais_2021_4751204} which represents 19,240 literary fictions drawn from Gallica, the extensive digitization initiative of the National Library of France. It is made up of works initially categorized as prose literary fiction spanning the period of 1600-1950. The collection's massive scale can be attributed to the efficiency of the French legal deposit system\footnote{\url{https://en.wikipedia.org/wiki/Legal_deposit\#France}}. This legal requirement mandates that publishers submit copies of all published works to the National Library. As a result, it is estimated that approximately 40\% \cite{langlais_2021_4751204} of all novels published in France during the 19th century are available in Gallica, based on the BNF catalog. Thus, it offers a representative sample of 19th and early 20th-century literature, encompassing various works such as lesser-known books and subgenres. Figure \ref{fig:corpus_bar} shows the time distribution of the corpus, pointing the large peak of production in the late nineteenth century, with almost 3.000 novels in the 1890s.

In France, texts generally enter the public domain 70 years after the death of their authors. As the crucial date of 1950 approaches, the corpus of novels significantly decreases in size. This is because works published after 1950 are still under copyright, and their full text is not available for use without permission from the rights holders. This limitation affects the overall size and diversity of the corpus, particularly for more recent literature.

The raw corpus contained several issues, such as the optical character recognition (OCR) quality, the presence of complete works from an author, or multiple publications of the same work. To address these issues, we removed all versions of complete works, as our focus was on individual texts. For novels with multiple editions, we selected the first publication to have the closest date associated with a text and its writing date (e.g., 6 editions of Hugo's \textit{Toilers of the Sea} from 1866 to 1894). This allowed us to maintain a more accurate representation of the texts in relation to their original publication dates.

\begin{figure}[!ht]
    \centering
    \includegraphics[width=10cm]{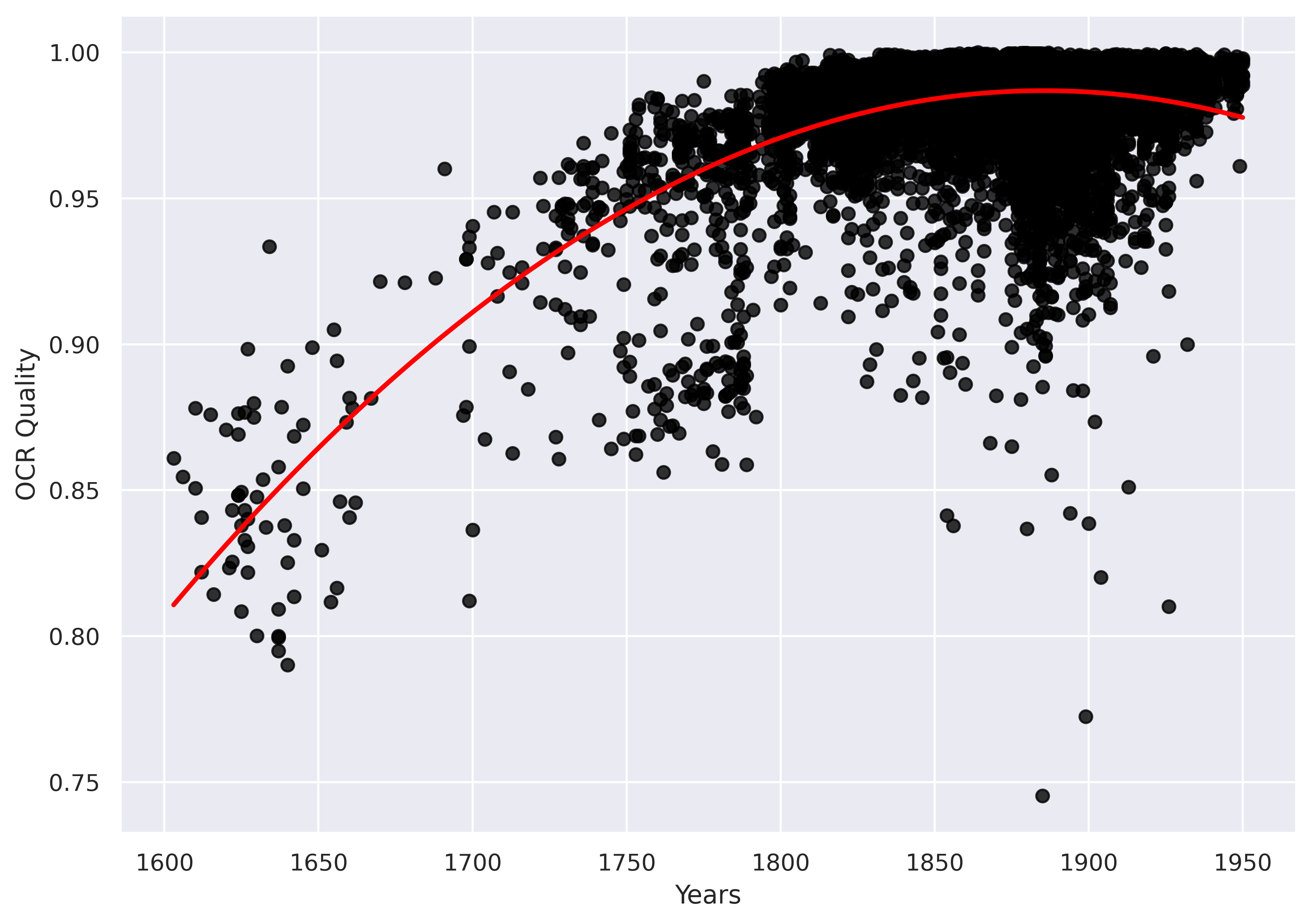}
    \caption{OCR quality over time}
    \label{fig:ocr_scat}
\end{figure}

Figure \ref{fig:ocr_scat} shows the temporal distribution of the works in the corpus based on a calculation of the proportion of correct words in each work. The goal is to detect works whose OCR quality is too poor for computational analysis. This evaluation of OCR quality was done by creating a dictionary of French words from a manually corrected set of literary texts. Each word in a given text was then compared to this dictionary, allowing us to compute a proxy for the word error rate metric.

As we can see, the texts before the beginning of the 19th century suffer from rather poor OCR quality or orthographic standardization. On the other hand, the decrease at the end of the century can be explained by a decline in the quality of paper used for certain works. This lower paper quality can negatively impact the OCR process, resulting in a lower overall text quality for these specific works. Filtering the corpus with an estimated OCR higher than 95\%, removing multiple publications and complete work from an author, we obtained a list of 12,176 novels. 

\subsection{Metadata construction}\label{metadata}


The notion of literary canon or the one of genre are not fixed and stable entities, but rather dynamic and contested constructions that reflects the values and ideologies of specific cultural and historic time. Therefore, it is important to approach these concepts with a critical and historical perspective, taking into account the complex and shifting dynamics of literary production and reception.

Setting aside the whole context of publication and reception of the work, which has an impact on textuality but cannot be recovered, the idea is to reintroduce historical context using proxies for literary reception and events that occurred during the life of the work. The challenge with this approach lies in the difficulty of finding large-scale viable metadata. To address this, we relied on previous research \cite{barre_operationalizing_2023} which focused on canonicity in the context of contemporary reception in French literature. We then applied it to our corpus, resulting in 11,103 non canonical elements and 1,073 canonical ones. 

For the genre labels, we decided to focus on a single subgenre, as we required clearly defined and coherent labels. We concentrated on “adventure novels" a dominant subgenre in the late nineteenth century. For this purpose, we relied on the work of Letourneux \cite{letourneux_roman_2010}, who defines a specific period (1870-1930) for the genre, allowing him to identify its key constants: “The importance of exotic settings [...] and the central role attributed to violent action, where the hero faces the risk of death or at least physical peril." We then generated synthetic metadata using a binary SVM classifier \cite{scikit-learn_2011}, trained on Letourneux's labels (with only 102 adventure labels retrieved), and as a result, we identified 2,114 "adventure novel" labels in the corpus under study. 

\subsection{Operationalizing textuality}\label{operationalization}

This article employs intertextuality as its theoretical foundation, drawing upon the work of structuralists who initially developed the concept. Our aim is to merge their theoretical perspectives with empirical experiments on an extensive collection of literary texts. It can be argued that computational literary studies inherently operationalize intertextuality through quantitative text comparisons. These comparisons may encompass various textual aspects, such as lexical, semantic, or thematic dimensions. However, a challenge arises from the absence of a clear definition of the specific textual components that intertextual theorists refer to.

Computational literary scholars have been striving to develop methods for comparing different texts. To achieve this, they have utilized various techniques to extract information from texts, such as topic-based (LDA, neural models), lexicon-based (BoW), or more semantic comparisons centered on entities or characters \cite{lasse_kohlmeyer_novel_2021}. These methods were selected due to their high interpretability, as basic machine learning techniques could determine topic or lexical importance for specific purposes. Nevertheless, certain textual dimensions, like word order or plot progression, have often been disregarded to obtain more interpretable features. Preventing this, the use of text embeddings has developed in the past few years, with methods such as Paragraph Vectors \cite{quoc_distributed_2014} applied to prestige inquiries \cite{van_cranenburgh_vector_2019} or genre clustering ones \cite{sobchuk2023computationalthematicscomparingalgorithms}. However, the “black box" nature of these methods has deterred some researchers. At the level of passages, embedding models have shown their strength in retrieving complex textual elements. The primary reason for using these methods is the uncertainty surrounding the specific information we want to retrieve from the texts. Some literary analysis tasks may involve different textual elements, formal or semantic dimensions, and the embedding model is supposedly encoding these informations in its latent space. 

In this study, we employ an encoder model based on a transformer language models, specifically the M3-Embedding dense model \cite{bge-m3}, an open-source model developed by the Beijing Academy of Artificial Intelligence. At the time of the experiment, this model led the multilingual MTEB evaluation benchmark\footnote{The MTEB benchmark stands for the Massive Text Embedding Benchmark. It is a large-scale evaluation framework designed to assess the performance of text embedding models across a wide variety of natural language processing (NLP) tasks. See paper \cite{muennighoff_mteb_2022} and url for more \url{https://huggingface.co/blog/mteb}}. We also chose this model because we could fine-tune it on French literary language, a crucial factor for our research since many encoder models may lack sufficient French language data in their training corpora, and particularly French literary fiction. This model features a 8192 tokens window, which is significantly larger than that of a more traditional BERT (512), allowing us to process a wide range of text lengths. This extended window size enables the model to capture more contextual information, making it more suitable for analyzing longer parts of literary works.

Consequently, we fine-tuned our model on a passage similarity task. We constructed the training corpora by selecting 400,000 random passages from our corpus. The “query" resulted in a paragraph of ten sentences. For the “positive" relation with the query, we used the ten subsequent sentences. As the “negative" relation, we chose ten random sentences from the entire corpus. The underlying assumption is that authors maintain a consistent language in their novels, especially when dealing with consecutive passages. We also replaced all proper names with a specific token \textit{“[PROPN]"} to prevent the encoder model to cluster passages only based on character names. Proper names of characters have been extracted using BookNLP-fr \cite{melanie-becquet_booknlp-fr_2024}, a state-of-the-art NER pipeline for literary entities. The model will iteratively enhance its performance on the author attribution task by leveraging various cues, including formal elements, thematic content, and stylistic features. As it undergoes training, it is expected to develop a deeper understanding of these characteristics, allowing it to more accurately identify the unique voice and style of different texts and authors.

After the fine-tuning, the goal was to infer a vector representation for each novel in our corpus. Then, a challenge arises, as our encoder can only process a context window of 8096 tokens, while a typical novel contains around 100,000 tokens. We implemented the following approach to handle this crucial step. For each novel in our corpus, we first randomly draw 100 passages of it, then we run our fine tuned encoder on each passage. Finally we take the mean embedding of all passages from a novel to represent it as a unique embedding. Thus we obtained 12,176 embeddings, one for each novel. As a distance metric, we opted for the cosine similarity, since it is widely used in the NLP and CLS fields. Previous work managed to show that the cosine similarity between pairs of word embeddings had a robust correlation with human similarity judgments \cite{SIP-124}. However, when applying cosine distance to embeddings drawn from contextualized language models like ours, previous research found that a small number of "rogue dimensions" had a disproportionate impact on the cosine similarity calculation \cite{timkey-van-schijndel-2021-bark}, \cite{mu2018allbutthetop}. To prevent this issue we ran a Standard Scaler normalization before running the cosine distance between every pair of novels from our corpus.

\section{Results}\label{results}
\subsection{Sanity check}\label{evaluation}

To validate our approach at each computational step, we implemented a sanity check using a subset of one thousand novels of our corpus. For each novel, we generated five distinct random representations by selecting random passages, computing their vector representations, and averaging the resulting vectors. This produced five mean embeddings for each novel, resulting in a total of 5000 embeddings. The goal of the sanity check is to ensure that, for any given embedding, the four most similar embeddings (based on cosine similarity) correspond to the other versions of the same novel. If the method fails to identify the four closest versions correctly, we apply a penalty to the accuracy score. This test verifies whether our process of randomizing passages, vectorizing them, and averaging the vectors maintains or not the integrity of each novel's representation.

\begin{figure}[!ht]
    \centering
    \includegraphics[width=10cm]{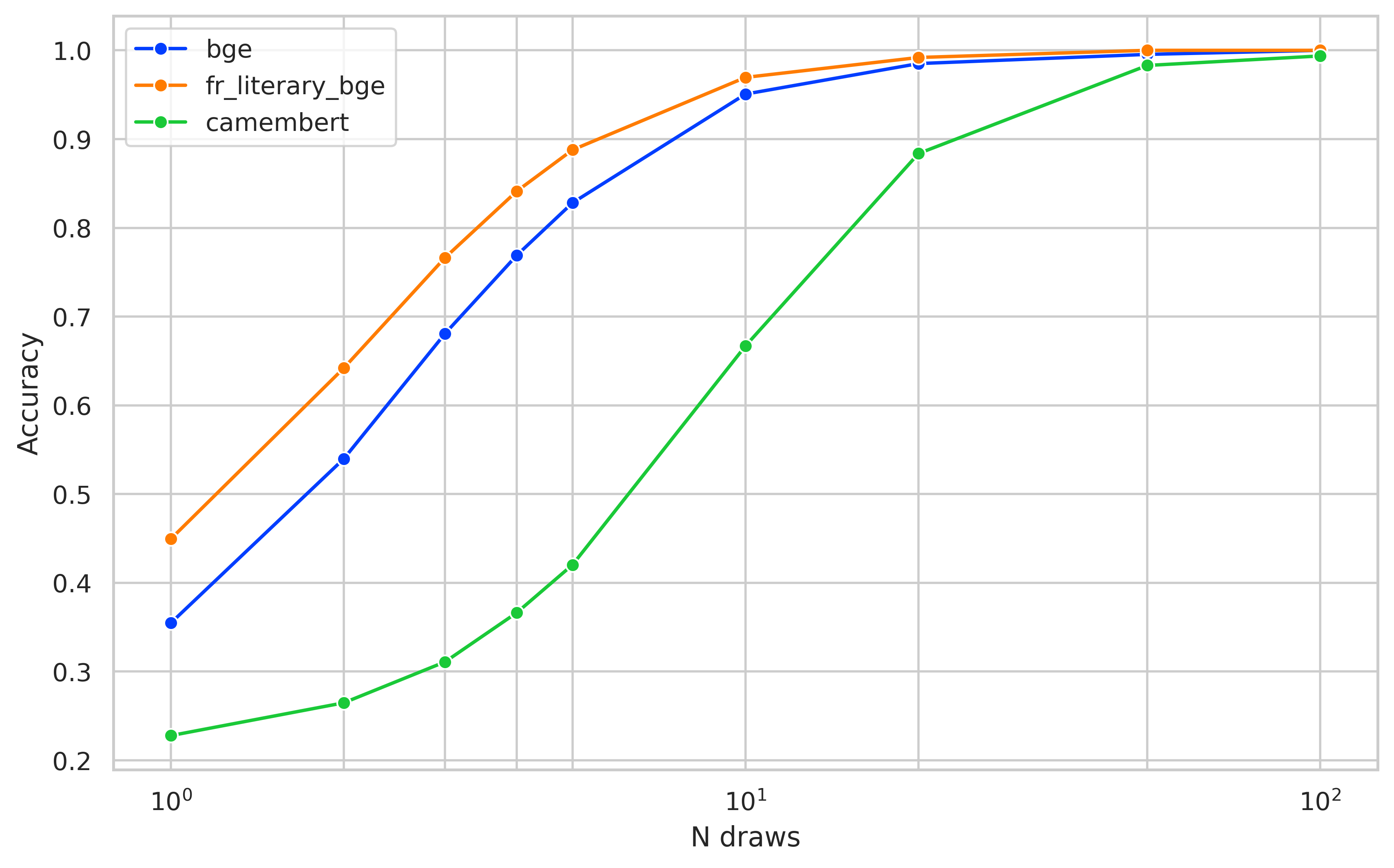}
    \caption{Sanity check accuracies between models and number of draws in novels}
    \label{fig:tirages}
\end{figure}

\autoref{fig:tirages} compares for each number of draws the performance of the base model BGE-M3, its fine-tuned version and the Camembert \cite{martin2020camembert} baseline. It shows that the fine-tuned model consistently outperforms the base model, and that the number of draws is halved to complete the sanity check (50 vs 100). This improvement highlights the effectiveness of fine-tuning, enabling an optimization of resource consumption at inference which is crucial when planning to launch the model on 12,000 novels.

Overall, these results suggest that our computational approach is effective at capturing the complexity of what constitutes a novel, while also being robust to the randomization of selected passages of the works. The evaluation we implemented turned out to be relatively easy for the model, as a 100\% accuracy was achieved quickly. However, this allowed us to determine the number of random draws needed to capture the complexity and uniqueness of a given novel.

\subsection{Similarity over time}\label{main_res}

\begin{figure}[!ht]
    \centering
    \includegraphics[width=10cm]{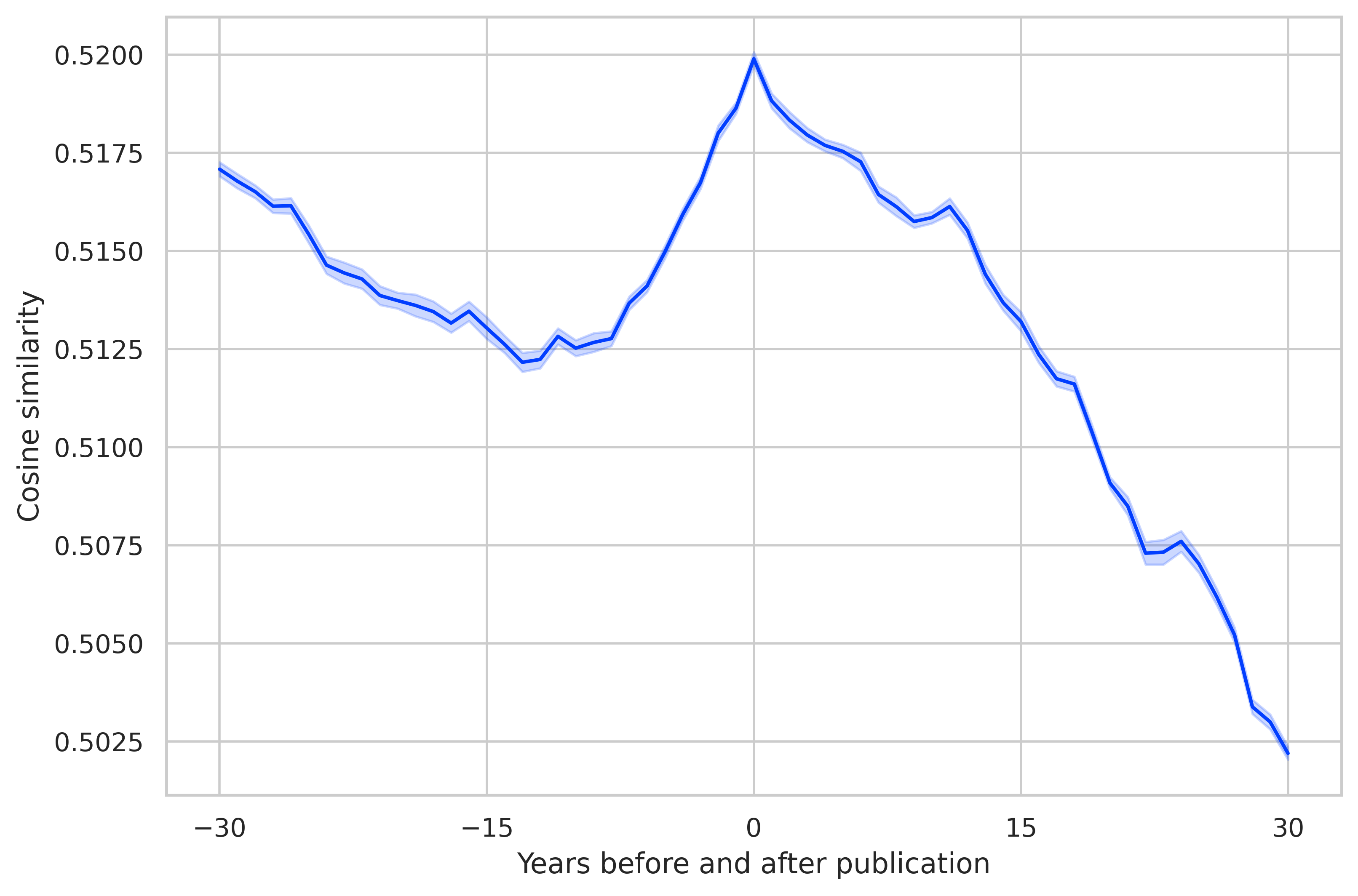}
    \caption{Similarity between texts considering dates before and after publication}
    \label{fig:sim-main}
\end{figure}

After retrieving the embeddings for each of our approximately 12,000 novels, a similarity matrix is constructed, with as many rows as columns, where we measure the cosine similarity of each text with all the others. On each row, works by the same author are set to NaN to prevent the authorial idiolect from skewing the temporal analyses. Next, a similarity matrix is computed for each text with every year in the corpus. To ensure statistical relevance, we apply downsampling to each year, meaning that every year must have a minimum of 25 novels and a maximum of 50 novels. Each text is then aligned with its publication year, centering the analysis at 0 for its date of publication, resulting in the graph presented in \autoref{fig:sim-main}. This downsampling process is repeated ten times, and the standard error is displayed in the graph. We chose 30 years before and after because previous research showed it was a good window for measuring literary change\footnote{See \cite{underwood_cohort_2022} and \citet{moretti_graphs_2007}'s work on genres, their analysis focused on the cycle of change, considering a timeframe of 25 to 30 years.}. A larger window would have also resulted in a significant loss of data in the analysis.

The resulting graph measures the similarity between each text and its context of production. A striking pattern emerges: a clear peak in similarity is observed during the year of the novel's publication. As the distance from the publication year increases, the textual similarity diminishes. This supports clearly previous research \cite{Hughes_2012}, where researchers provided quantitative evidence for the concept of a literary “style of a time", highlighting a strong trend toward more contemporary stylistic influences.

Surprisingly, before the similarity peak at the time of publication, there is a noticeable downward trend in the cosine similarity between -30 and -15 years before publication. The high similarity 30 years before publication suggests that earlier works had a strong influence on the author's development. This influence decreases as the publication date approaches, reflecting the gradual divergence from older literary models. The trend aligns with previous research on the cohort effect, where an author’s writing is shaped by books from their formative years. However, this goes beyond the analysis of this paper, and we have no evidence to support this hypothesis here. \citet{underwood_cohort_2022} provided strong evidence of cohorts driving literary change, though their work did not specifically explore their effect on textual similarity.

\subsection{Individual similarity over time}\label{indiv_res}

\begin{figure}[!ht]
    \centering
    \begin{subfigure}[b]{0.45\textwidth}
        \centering
        \includegraphics[width=\textwidth]{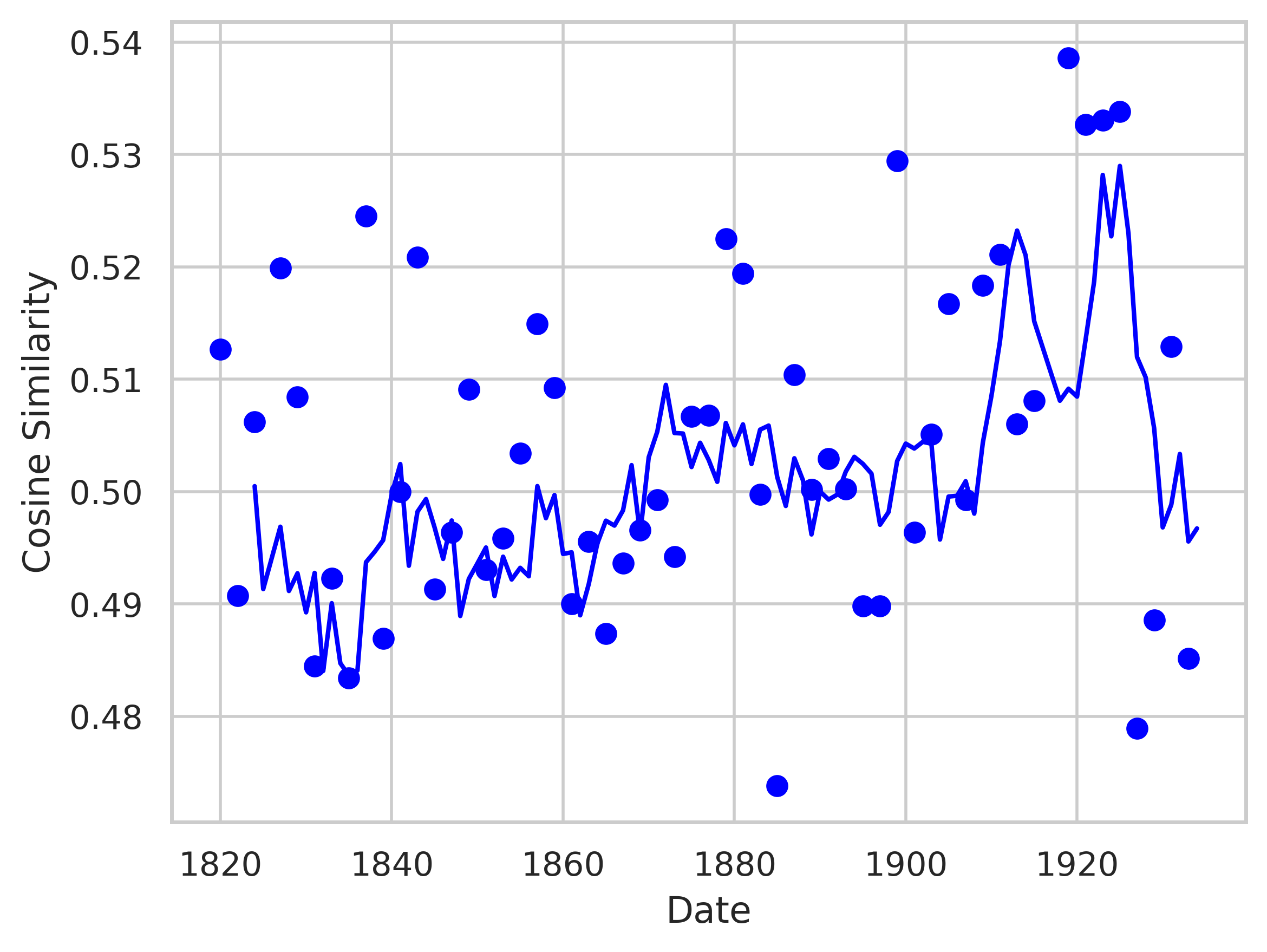}
        \caption{Cosine similarity over time for Proust's \textit{Le temps retrouvé}}
        \label{fig:plot1}
    \end{subfigure}
    \hfill
    \begin{subfigure}[b]{0.45\textwidth}
        \centering
        \includegraphics[width=\textwidth]{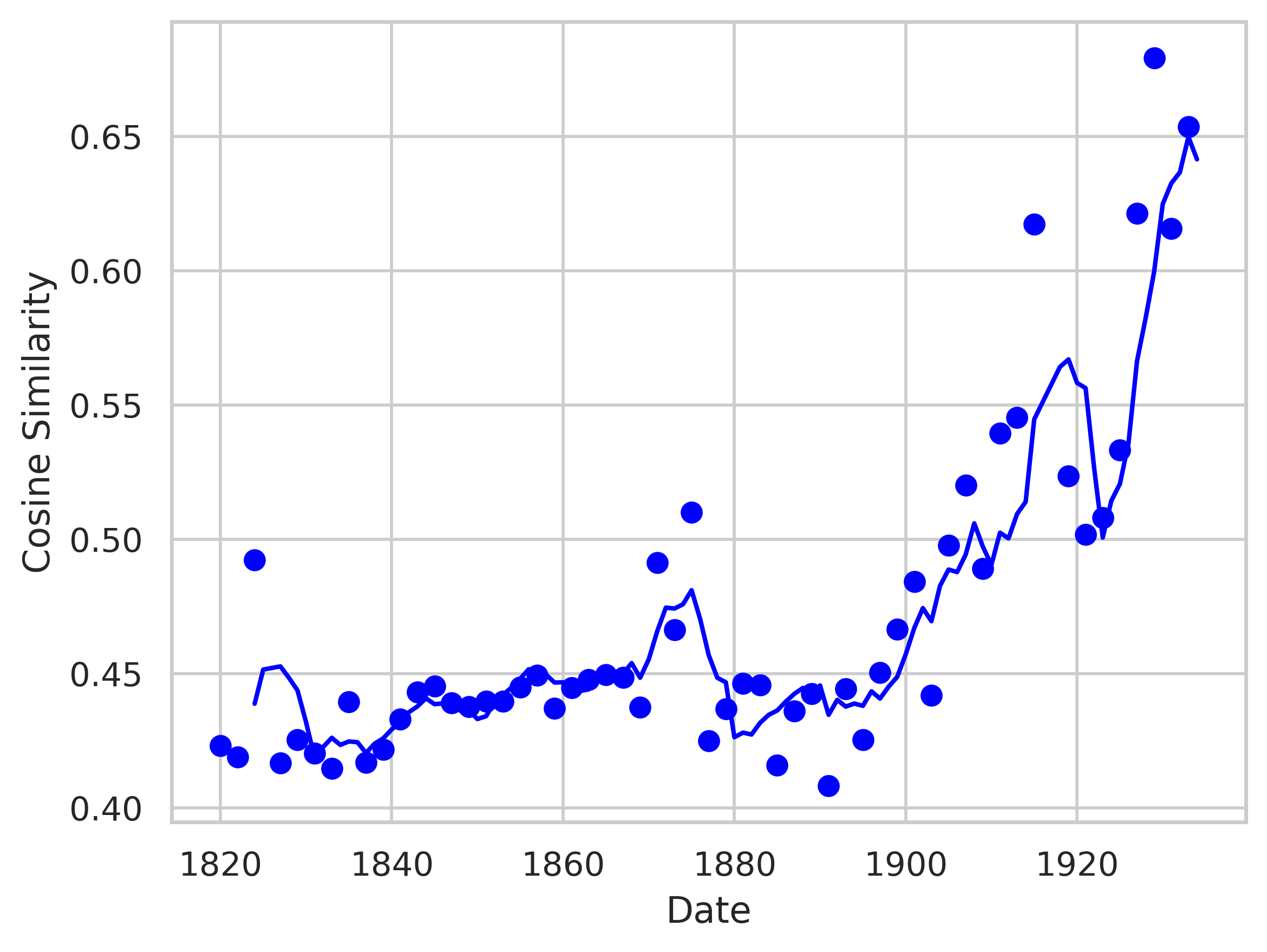}
        \caption{Cosine similarity over time for Leroux's \textit{Le mystère de la chambre jaune}}
        \label{fig:plot2}
    \end{subfigure}
    
    \vspace{1em} 

    \begin{subfigure}[b]{0.45\textwidth}
        \centering
        \includegraphics[width=\textwidth]{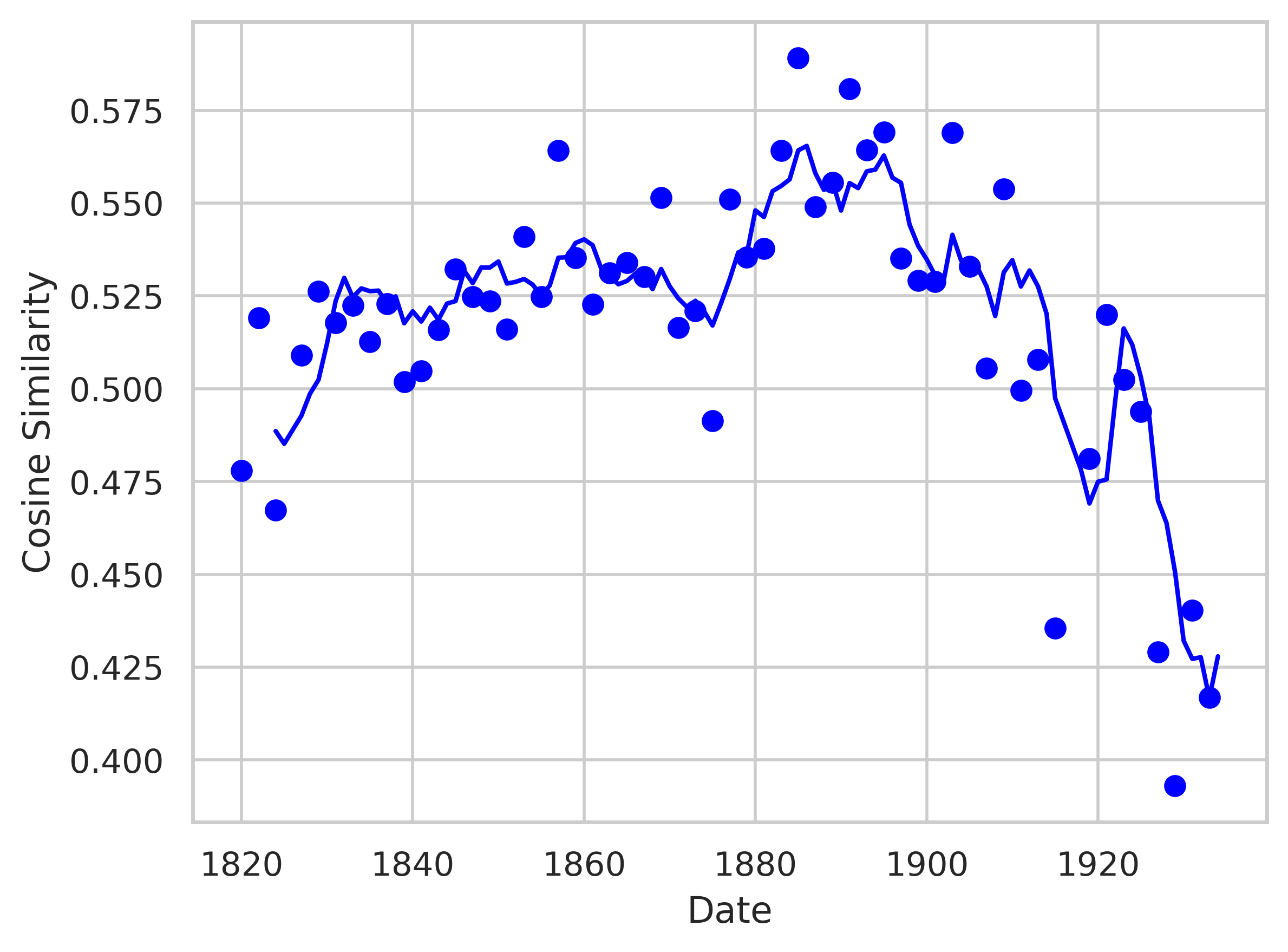}
        \caption{Cosine similarity over time for Hugo's \textit{Les Misérables}}
        \label{fig:plot3}
    \end{subfigure}
    \hfill
    \begin{subfigure}[b]{0.45\textwidth}
        \centering
        \includegraphics[width=\textwidth]{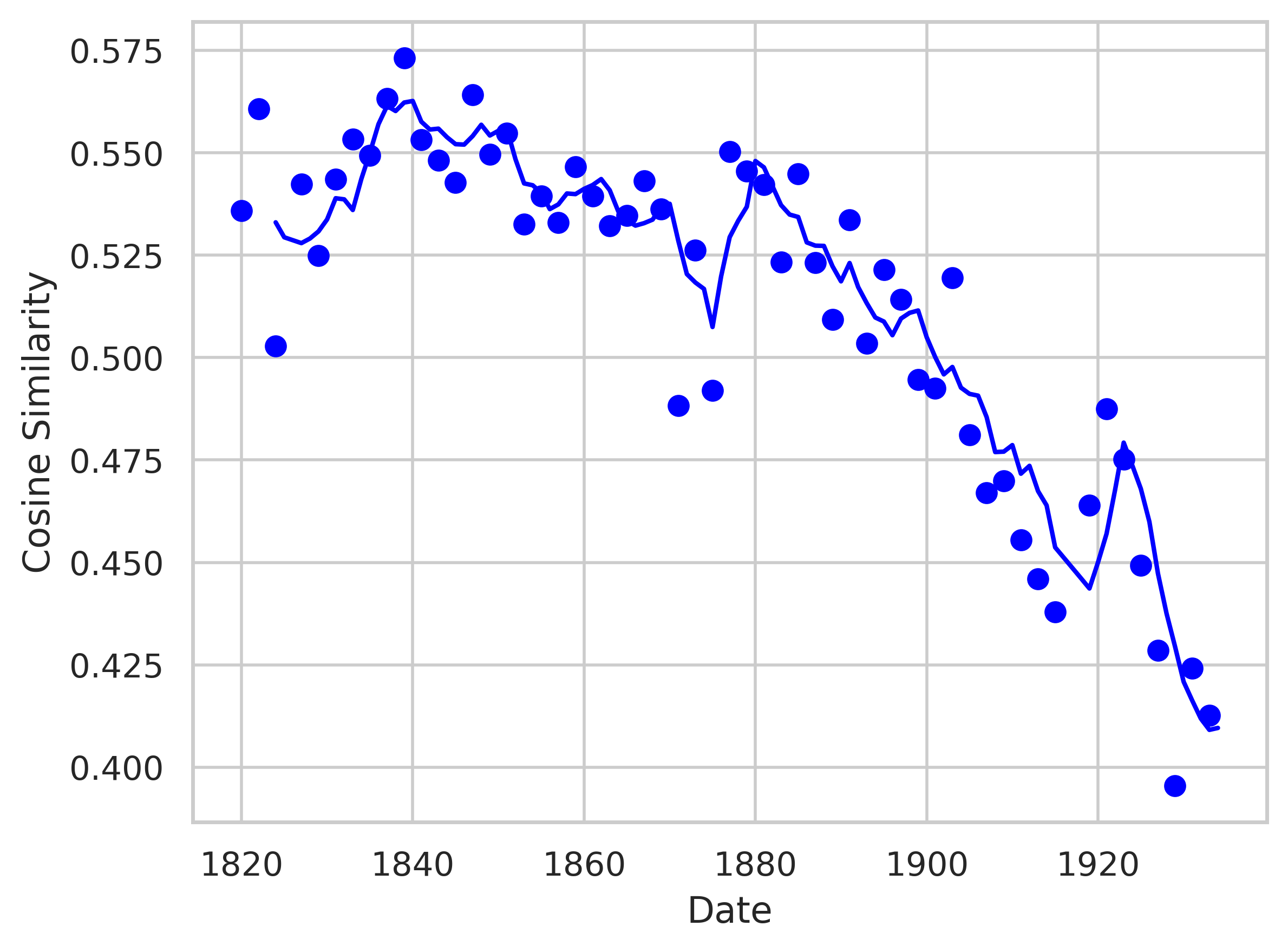}
        \caption{Cosine similarity over time for Reybaud's \textit{Mézélie}}
        \label{fig:plot4}
    \end{subfigure}
    
    \caption{Different patterns of individual textual similarity evolution over time}
    \label{fig:all_plots}
\end{figure}

Then, we wanted to grasp the individual trends of specific novels to investigate how they move in the textual similarity network. For these experiments, we took back the downsampled similarity matrix computed for each text with every year in the corpus. \autoref{fig:all_plots} represents four different patterns of textual similarity over time. \autoref{fig:plot1} shows the cosine similarity plot for Proust's \textit{Le temps retrouvé}, the final volume of his monumental \textit{À la recherche du temps perdu}. The graph reveals a significant peak around the time of its post-mortem publication in 1927. The detected pattern is not the author's signature, as texts by the same author are excluded from the analysis. Instead, it reflects a kind of "average language" or a broader intertextual network at the linguistic level. This peak aligns deeply with the modernist focus on subjectivity and inner consciousness. Like many modernist writers of the time, Proust was influenced by emerging psychological theories, especially those of Sigmund Freud, which emphasized the exploration of the unconscious mind. In \textit{Le temps retrouvé}, Proust explores the protagonist’s memories, perceptions, and internal reflections, creating a narrative that is more concerned with the flow of time as experienced subjectively than with external events or linear plot progression. Proust's plot exhibits the highest level of noise among the four. This can be explained by insights from \citet{compagnon_proust_2013}'s work which provides a framework for understanding how Proust's \textit{À la recherche du temps perdu} reflects a dialogue between past literary forms and the new modernist literary movement that was gaining prominence in the early 20th century. This concept of Proust being “between two centuries" helps explain why Proust's work resonates with both 19th-century novelists and modernist writers, which could explain the disparities in similarity across time.

\autoref{fig:plot2} shows the similarity over time of Leroux’s \textit{Le mystère de la chambre jaune}, published in 1907. A huge increase in cosine similarity appears from the beginning of the twentieth century to the end of the observed time period. This reflects perfectly the novel's status as a “proto-detective" story \cite{lavergne_naissance_2009} that laid the groundwork for the \textit{whodunit} detective genre. This peak can be attributed to the rise of detective fiction following Leroux’s or Doyle's innovations, with authors building on their narrative techniques and conventions. As detective stories proliferated in the early 20th century, the post-publication similarity of \textit{Le mystère de la chambre jaune} became increasingly apparent.

In \autoref{fig:plot3}, Victor Hugo’s \textit{Les Misérables} offers a more intriguing pattern. If a slight peak around its publication date in 1862 appears, it is much less pronounced than the decades that follow between 1880 and 1900. This could be explained by the growing canonical status of both Hugo and \textit{Les Misérables} influencing subsequent generations of writers who either drew inspiration from its themes or echoed its narrative structures. The vivid depictions of social injustice, poverty, and the struggles of the lower classes in \textit{Les Misérables} resonate with key themes of Naturalism, despite the fact that they will take this social critique further, focusing on deterministic views of human behavior influenced by environment and heredity. The drop at the end of the graph is likely due to linguistic and thematic changes that are too significant to maintain high textual similarity.

The last individual trend in \autoref{fig:plot4} is \textit{Mézélie}, published in 1839 by Madame Charles Reybaud. The graph exhibits a gradual decline in cosine similarity over time. Despite her notable success and popularity during her time, Reybaud's works, including \textit{Mézélie}, have largely faded into obscurity and is now part of the so called “archive".

\subsection{Collective Structuring Frames of Textual Similarity}\label{collective_res}

Our final experiments use the same type of analysis as in \autoref{main_res}, but we aimed to explore the extent to which collective factors, such as genres or canonicity, could influence the different trends in textual observed earlier. To that end we relied on metadata described in \autoref{metadata}. To compare our samples (canon vs archive or adventure vs general) with equal temporal cardinality, we implemented stratified sampling. We randomly selected the appropriate number of elements from the general and non-canon samples to match those in the adventure and canonical sets, while also maintaining the temporal distribution per decades of these specific groups.

\begin{figure}[!ht]
    \centering
    \begin{subfigure}[b]{0.45\textwidth}
        \centering
        \includegraphics[width=\textwidth]{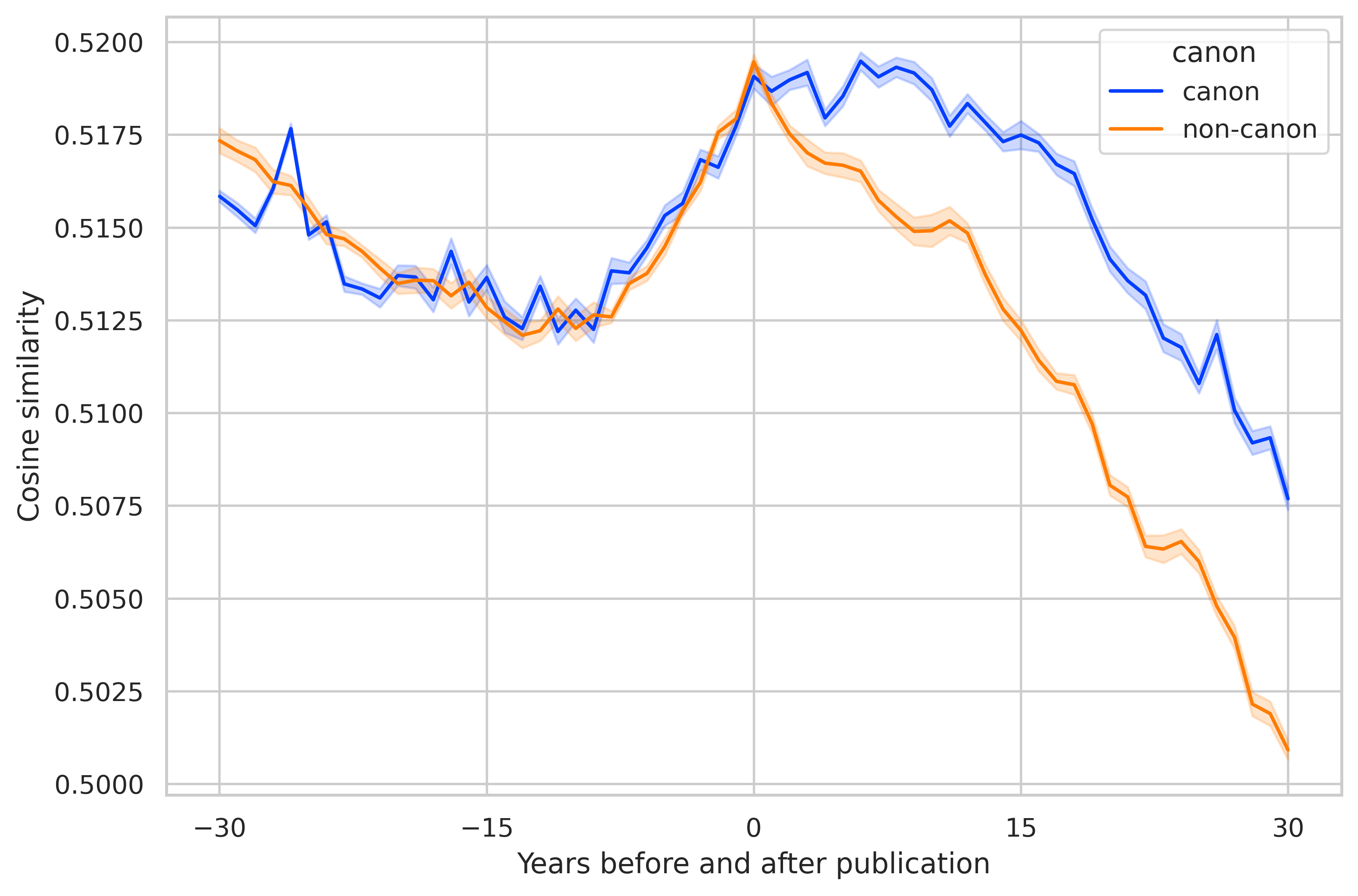}
        \caption{Similarity between texts considering dates before and after publication, Canon in blue, Archive in orange}
        \label{fig:sim-canon}
    \end{subfigure}
    \hfill
    \begin{subfigure}[b]{0.45\textwidth}
        \centering
        \includegraphics[width=\textwidth]{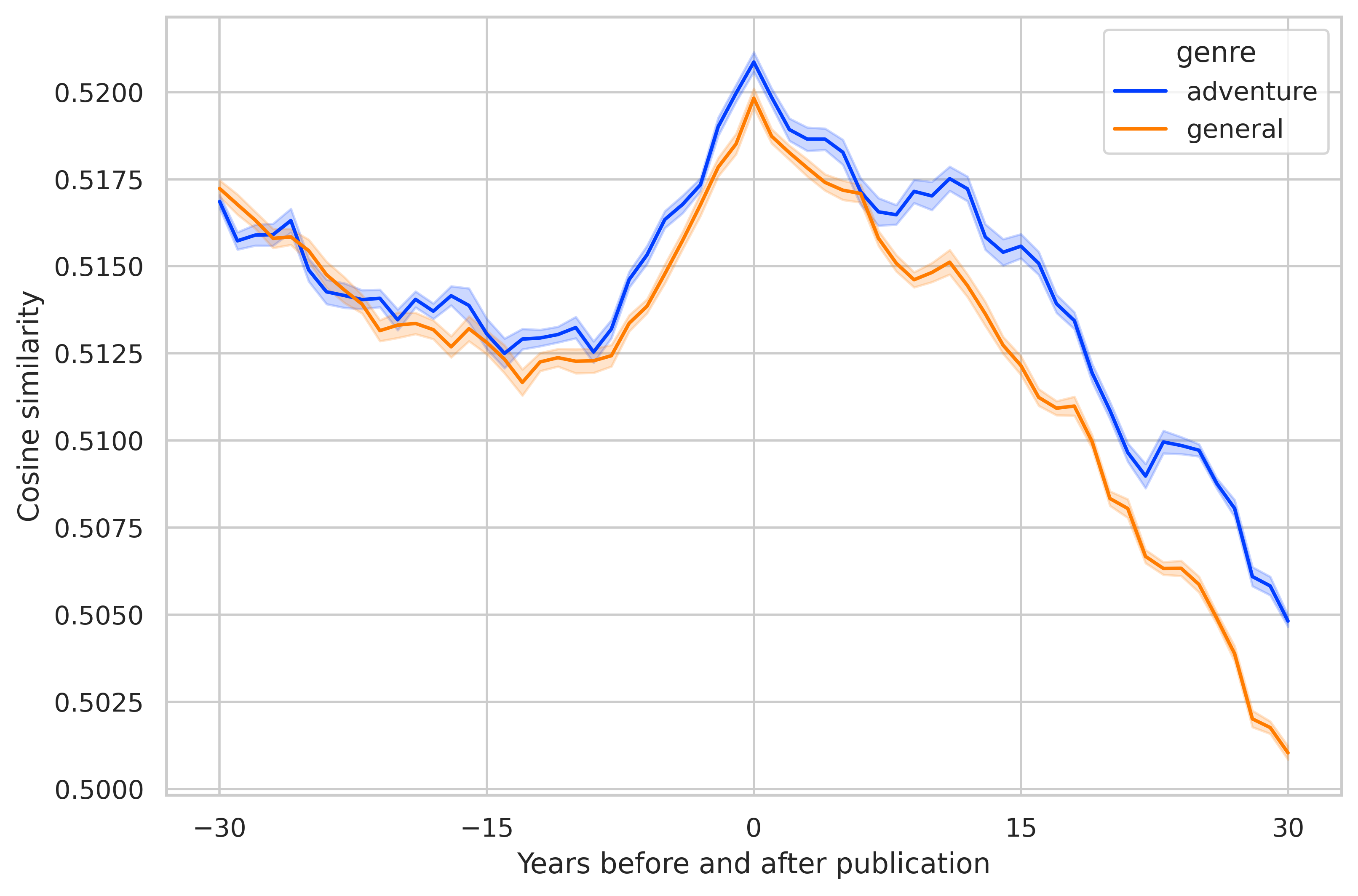}
        \caption{Similarity between texts considering dates before and after publication, Adventure in blue,  in orange}
        \label{fig:sim-aventures}
    \end{subfigure}
    \caption{Different patterns of collective textual similarity evolution over time}
\end{figure}

\autoref{fig:sim-canon} shows that while both canonical and archival novels follow a similar similarity trend before publication, the curves diverge notably after the publication date. The pattern is particularly striking, suggesting that canonical works tend to exhibit higher similarity scores with later publications compared to archival works. This trend persists over time, indicating that canonical works exert a more enduring influence on the intertextual network, whereas non-canonical works have a more limited and shorter-lived impact.

\autoref{fig:sim-aventures} shows a less striking pattern but remains interesting and statistically solid. Both curves follow a similar trend before publication, yet they begin to diverge approximately 10 years before and up to 5 years after the publication date. This pattern suggests that genres have a precise contextual existence within a specific historical moment. Following the decline in similarity, we observe that adventure novels maintain a stronger similarity with later texts (from +10 to +30 years). This is harder to interpret, but one hypothesis could be that many works within the genre were canonized retrospectively, which, as seen in \autoref{fig:sim-canon} results in a higher similarity with subsequent works.

\section{Discussion}\label{discussion}



In this article, we demonstrated that both subgenres and canonicity function as collective structuring frames for textuality. While our operationalization using language model encoders and cosine similarity is undoubtedly limited in its ability to fully capture the complexity of a novel, our approach nonetheless uncovered novel patterns of similarity within French fiction.

Firstly, we showed that canonical works tend to be more deeply integrated into the intertextual network after their publication. This investigation sheds light on the collective representations that shape the cultural frameworks within which we operate. One way to understand this phenomenon is through the concept of “cultural grammar," as proposed by \citet{altieri_idea_1983}. According to this view, canonical literary works serve as foundational texts that help to shape the norms, values, and conventions within a particular cultural tradition.

Secondly, we highlighted that texts within the same genre share distinct textual similarities during the relatively brief period when the genre is being defined. This could be attributed to the commercial nature of genre, which contributes to increasing the similarity of certain texts within a specific historical moment.

Ultimately, both canonical and subgenres works establish a set of shared references and expectations that guide the production and reception of subsequent texts, and contribute to the formation of a shared cultural imagination. 

Narrowing the scope of study appears essential for future research in two ways: reducing the number of works analyzed and focusing on passages rather than entire novels. This will help to reestablish a more concrete foundation based on textual evidence. It would be interesting for example, to examine these dynamics at the scale of a specific sub-genre, taking the example of detective novels in order to track the dynamics of intertextuality within a highly codified and well-defined sub-genre. This will also provide an opportunity to test the distant reading observations on the subgenre's canonical novels, to determine whether the phenomenon persists at a smaller scale.

\begin{acknowledgments}
Jean Barré’s PhD is supported by the EUR (Ecole Universitaire de Recherche) Translitteræ (programme “Investissements d’avenir” ANR-10- IDEX-0001-02 PSL and ANR-17-EURE-0025).
\end{acknowledgments}

\bibliography{main}

\begin{thebibliography}{37}
\expandafter\ifx\csname natexlab\endcsname\relax\def\natexlab#1{#1}\fi
\providecommand{\url}[1]{\texttt{#1}}
\providecommand{\href}[2]{#2}
\providecommand{\path}[1]{#1}
\providecommand{\DOIprefix}{doi:}
\providecommand{\ArXivprefix}{arXiv:}
\providecommand{\URLprefix}{URL: }
\providecommand{\Pubmedprefix}{pmid:}
\providecommand{\doi}[1]{\href{http://dx.doi.org/#1}{\path{#1}}}
\providecommand{\Pubmed}[1]{\href{pmid:#1}{\path{#1}}}
\providecommand{\bibinfo}[2]{#2}
\ifx\xfnm\relax \def\xfnm[#1]{\unskip,\space#1}\fi
\bibitem[{Hughes et~al.(2012)Hughes, Foti, Krakauer, and Rockmore}]{Hughes_2012}
\bibinfo{author}{J.~M. Hughes}, \bibinfo{author}{N.~J. Foti}, \bibinfo{author}{D.~C. Krakauer}, \bibinfo{author}{D.~N. Rockmore},
\newblock \bibinfo{title}{Quantitative patterns of stylistic influence in the evolution of literature},
\newblock \bibinfo{journal}{Proceedings of the National Academy of Sciences} \bibinfo{volume}{109} (\bibinfo{year}{2012}) \bibinfo{pages}{7682–7686}. \DOIprefix\doi{10.1073/pnas.1115407109}.
\bibitem[{Kristeva(2017)}]{kristeva_semeiotike_2017}
\bibinfo{author}{J.~Kristeva}, \bibinfo{title}{Sèméiotikè: recherches pour une sémanalyse}, Points, \bibinfo{publisher}{Éditions Point}, \bibinfo{address}{Paris}, \bibinfo{year}{2017}.
\bibitem[{Barthes(1974)}]{Barthes_1974}
\bibinfo{author}{R.~Barthes},
\newblock \bibinfo{title}{Texte (théorie du)},
\newblock \bibinfo{journal}{Encyclopædia Universalis}  (\bibinfo{year}{1974}). \URLprefix \url{https://www.universalis-edu.com/encyclopedie/theorie-du-texte/}.
\bibitem[{Allen(2022)}]{allen_intertextuality_2022}
\bibinfo{author}{G.~Allen}, \bibinfo{title}{Intertextuality}, The new critical idiom, \bibinfo{edition}{third edition} ed., \bibinfo{publisher}{Routledge, Taylor \& Francis Group}, \bibinfo{year}{2022}.
\bibitem[{Allen et~al.(2010)Allen, Cooney, Douard, Horton, Morrisse, Olsen, Roe, and Voyer}]{allen_plundering_2010}
\bibinfo{author}{T.~Allen}, \bibinfo{author}{C.~Cooney}, \bibinfo{author}{S.~Douard}, \bibinfo{author}{R.~Horton}, \bibinfo{author}{R.~Morrisse}, \bibinfo{author}{M.~Olsen}, \bibinfo{author}{G.~Roe}, \bibinfo{author}{R.~Voyer},
\newblock \bibinfo{title}{Plundering {Philosophers}: {Identifying} {Sources} of the {Encyclopédie}},
\newblock \bibinfo{journal}{Journal of the Association for History and Computing}  (\bibinfo{year}{2010}). \URLprefix \url{http://hdl.handle.net/2027/spo.3310410.0013.107}.
\bibitem[{Büchler et~al.(2012)Büchler, Crane, Moritz, and Babeu}]{B_chler_2012}
\bibinfo{author}{M.~Büchler}, \bibinfo{author}{G.~Crane}, \bibinfo{author}{M.~Moritz}, \bibinfo{author}{A.~Babeu}, \bibinfo{title}{Increasing Recall for Text Re-use in Historical Documents to Support Research in the Humanities}, \bibinfo{publisher}{Springer Berlin Heidelberg}, \bibinfo{year}{2012}, p. \bibinfo{pages}{95–100}. \DOIprefix\doi{10.1007/978-3-642-33290-6_11}.
\bibitem[{Ganascia et~al.(2014)Ganascia, Glaudes, and Del~Lungo}]{ganascia_2014}
\bibinfo{author}{J.-G. Ganascia}, \bibinfo{author}{P.~Glaudes}, \bibinfo{author}{A.~Del~Lungo}, \bibinfo{title}{Automatic detection of reuses and citations in literary texts}, \bibinfo{year}{2014}. \DOIprefix\doi{10.48550/ARXIV.1404.2997}.
\bibitem[{Ganascia(2020)}]{Ganascia_2020}
\bibinfo{author}{J.-G. Ganascia},
\newblock \bibinfo{title}{Détection automatique de phénomènes intertextuels},
\newblock \bibinfo{journal}{Genesis}  (\bibinfo{year}{2020}) \bibinfo{pages}{63–77}. \DOIprefix\doi{10.4000/genesis.5671}.
\bibitem[{Manjavacas et~al.(2020)Manjavacas, Karsdorp, and Kestemont}]{manjavacas_statistical_2020}
\bibinfo{author}{E.~Manjavacas}, \bibinfo{author}{F.~Karsdorp}, \bibinfo{author}{M.~Kestemont}, \bibinfo{title}{A Statistical Foray into Contextual Aspects of Intertextuality}, volume \bibinfo{volume}{2723}, \bibinfo{publisher}{CEUR Workshop Proceedings}, \bibinfo{year}{2020}, pp. \bibinfo{pages}{77--96}.
\bibitem[{Algee-Hewitt et~al.(2016)Algee-Hewitt, Allison, Gemma, Heuser, Walser, and Moretti}]{algee-hewitt_canonarchive_2016}
\bibinfo{author}{M.~Algee-Hewitt}, \bibinfo{author}{S.~Allison}, \bibinfo{author}{M.~Gemma}, \bibinfo{author}{R.~Heuser}, \bibinfo{author}{H.~Walser}, \bibinfo{author}{F.~Moretti},
\newblock \bibinfo{title}{Canon/archive. large-scale dynamics in the literary field},
\newblock \bibinfo{journal}{Pamphlets of the Stanford Literary Lab}  (\bibinfo{year}{2016}). \URLprefix \url{https://litlab.stanford.edu/LiteraryLabPamphlet11.pdf}.
\bibitem[{Underwood(2019)}]{underwood_distant_2019}
\bibinfo{author}{T.~Underwood}, \bibinfo{title}{Distant horizons: digital evidence and literary change}, \bibinfo{publisher}{The University of Chicago Press}, \bibinfo{year}{2019}.
\bibitem[{Brottrager et~al.(2022)Brottrager, Stahl, Arslan, Brandes, and Weitin}]{judith_brottrager_modeling_2022}
\bibinfo{author}{J.~Brottrager}, \bibinfo{author}{A.~Stahl}, \bibinfo{author}{A.~Arslan}, \bibinfo{author}{U.~Brandes}, \bibinfo{author}{T.~Weitin},
\newblock \bibinfo{title}{Modeling and predicting literary reception. a data-rich approach to literary historical reception},
\newblock \bibinfo{journal}{Journal of Computational Literary Studies} \bibinfo{volume}{1} (\bibinfo{year}{2022}). \DOIprefix\doi{10.48694/jcls.95}, \bibinfo{note}{publisher: Universitäts- und Landesbibliothek Darmstadt}.
\bibitem[{Barré et~al.(2023)Barré, Camps, and Poibeau}]{barre_operationalizing_2023}
\bibinfo{author}{J.~Barré}, \bibinfo{author}{J.-B. Camps}, \bibinfo{author}{T.~Poibeau},
\newblock \bibinfo{title}{Operationalizing canonicity: A quantitative study of french 19th and 20th century literature},
\newblock \bibinfo{journal}{Journal of Cultural Analytics} \bibinfo{volume}{8} (\bibinfo{year}{2023}). \DOIprefix\doi{10.22148/001c.88113}.
\bibitem[{Barré and Poibeau(2023)}]{barre_beyond_2023}
\bibinfo{author}{J.~Barré}, \bibinfo{author}{T.~Poibeau},
\newblock \bibinfo{title}{Beyond canonicity: Modeling canon/archive literary change in french fiction},
\newblock in: \bibinfo{booktitle}{CEUR Workshop Proceedings CHR2023}, \bibinfo{year}{2023}, pp. \bibinfo{pages}{814--830}.
\bibitem[{Altieri(1983)}]{altieri_idea_1983}
\bibinfo{author}{C.~Altieri},
\newblock \bibinfo{title}{An idea and ideal of a literary canon},
\newblock \bibinfo{journal}{Critical Inquiry} \bibinfo{volume}{10} (\bibinfo{year}{1983}) \bibinfo{pages}{37--60}. \DOIprefix\doi{10.1086/448236}.
\bibitem[{Genette(1986)}]{genette_theorie_1986}
\bibinfo{author}{G.~Genette},
\newblock \bibinfo{title}{Introduction à l'architexte},
\newblock in: \bibinfo{editor}{G.~Genette}, \bibinfo{editor}{T.~Todorov} (Eds.), \bibinfo{booktitle}{Théorie des genres}, number \bibinfo{number}{181} in \bibinfo{series}{Points}, \bibinfo{publisher}{Éd. du Seuil}, \bibinfo{year}{1986}, pp. \bibinfo{pages}{110--148}.
\bibitem[{Barr{\'e}(2024)}]{barre:hal-04559749}
\bibinfo{author}{J.~Barr{\'e}},
\newblock \bibinfo{title}{{D{\'e}tection automatique de l'architextualit{\'e} dans le roman d'aventures}},
\newblock in: \bibinfo{booktitle}{{Humanistica 2024}}, Stylom{\'e}trie, \bibinfo{organization}{{Association francophone des humanit{\'e}s num{\'e}riques}}, \bibinfo{address}{Mekn{\`e}s, Morocco}, \bibinfo{year}{2024}. \URLprefix \url{https://hal.science/hal-04559749}.
\bibitem[{Jauß(2010)}]{jaus_toward_2010}
\bibinfo{author}{H.~R. Jauß}, \bibinfo{title}{Toward an aesthetic of reception}, number~\bibinfo{number}{2} in \bibinfo{series}{Theory and history of literature}, \bibinfo{edition}{nachdr.} ed., \bibinfo{publisher}{Univ. of Minnesota Press}, \bibinfo{year}{2010}.
\bibitem[{Langlais(2021)}]{langlais_2021_4751204}
\bibinfo{author}{P.-C. Langlais}, \bibinfo{title}{{Fictions littéraires de Gallica / Literary fictions of Gallica}}, \bibinfo{year}{2021}. \DOIprefix\doi{10.5281/zenodo.4751204}.
\bibitem[{Letourneux(2010)}]{letourneux_roman_2010}
\bibinfo{author}{M.~Letourneux}, \bibinfo{title}{Le roman d'aventures: 1870-1930}, \bibinfo{publisher}{Presses Universitaires de Limoges et du Limousin}, \bibinfo{year}{2010}.
\bibitem[{Pedregosa et~al.(2011)Pedregosa, Varoquaux, Gramfort, Michel, Thirion, Grisel, Blondel, Prettenhofer, Weiss, Dubourg, Vanderplas, Passos, Cournapeau, Brucher, Perrot, and Duchesnay}]{scikit-learn_2011}
\bibinfo{author}{F.~Pedregosa}, \bibinfo{author}{G.~Varoquaux}, \bibinfo{author}{A.~Gramfort}, \bibinfo{author}{V.~Michel}, \bibinfo{author}{B.~Thirion}, \bibinfo{author}{O.~Grisel}, \bibinfo{author}{M.~Blondel}, \bibinfo{author}{P.~Prettenhofer}, \bibinfo{author}{R.~Weiss}, \bibinfo{author}{V.~Dubourg}, \bibinfo{author}{J.~Vanderplas}, \bibinfo{author}{A.~Passos}, \bibinfo{author}{D.~Cournapeau}, \bibinfo{author}{M.~Brucher}, \bibinfo{author}{M.~Perrot}, \bibinfo{author}{E.~Duchesnay},
\newblock \bibinfo{title}{Scikit-learn: Machine learning in {P}ython},
\newblock \bibinfo{journal}{Journal of Machine Learning Research} \bibinfo{volume}{12} (\bibinfo{year}{2011}) \bibinfo{pages}{2825--2830}.
\bibitem[{Kohlmeyer et~al.(2021)Kohlmeyer, Repke, and Krestel}]{lasse_kohlmeyer_novel_2021}
\bibinfo{author}{L.~Kohlmeyer}, \bibinfo{author}{T.~Repke}, \bibinfo{author}{R.~Krestel},
\newblock \bibinfo{title}{Novel views on novels: Embedding multiple facets of long texts},
\newblock \bibinfo{journal}{2021 Association for Computing Machinery.}  (\bibinfo{year}{2021}).
\bibitem[{Le and Mikolov(2014)}]{quoc_distributed_2014}
\bibinfo{author}{Q.~V. Le}, \bibinfo{author}{T.~Mikolov}, \bibinfo{title}{Distributed representations of sentences and documents}, \bibinfo{year}{2014}. \URLprefix \url{https://arxiv.org/abs/1405.4053}. \href{http://arxiv.org/abs/1405.4053}{{\tt arXiv:1405.4053}}.
\bibitem[{van Cranenburgh et~al.(2019)van Cranenburgh, van Dalen-Oskam, and van Zundert}]{van_cranenburgh_vector_2019}
\bibinfo{author}{A.~van Cranenburgh}, \bibinfo{author}{K.~van Dalen-Oskam}, \bibinfo{author}{J.~van Zundert},
\newblock \bibinfo{title}{Vector space explorations of literary language},
\newblock \bibinfo{journal}{Language Resources and Evaluation} \bibinfo{volume}{53} (\bibinfo{year}{2019}) \bibinfo{pages}{625--650}. \DOIprefix\doi{10.1007/s10579-018-09442-4}.
\bibitem[{Sobchuk and Šeļa(2023)}]{sobchuk2023computationalthematicscomparingalgorithms}
\bibinfo{author}{O.~Sobchuk}, \bibinfo{author}{A.~Šeļa}, \bibinfo{title}{Computational thematics: Comparing algorithms for clustering the genres of literary fiction}, \bibinfo{year}{2023}. \href{http://arxiv.org/abs/2305.11251}{{\tt arXiv:2305.11251}}.
\bibitem[{Chen et~al.(2024)Chen, Xiao, Zhang, Luo, Lian, and Liu}]{bge-m3}
\bibinfo{author}{J.~Chen}, \bibinfo{author}{S.~Xiao}, \bibinfo{author}{P.~Zhang}, \bibinfo{author}{K.~Luo}, \bibinfo{author}{D.~Lian}, \bibinfo{author}{Z.~Liu}, \bibinfo{title}{M3-embedding: Multi-lingual, multi-functionality, multi-granularity text embeddings through self-knowledge distillation}, \bibinfo{year}{2024}. \href{http://arxiv.org/abs/2402.03216}{{\tt arXiv:2402.03216}}.
\bibitem[{Muennighoff et~al.(2022)Muennighoff, Tazi, Magne, and Reimers}]{muennighoff_mteb_2022}
\bibinfo{author}{N.~Muennighoff}, \bibinfo{author}{N.~Tazi}, \bibinfo{author}{L.~Magne}, \bibinfo{author}{N.~Reimers}, \bibinfo{title}{{MTEB}: {Massive} {Text} {Embedding} {Benchmark}}, \bibinfo{year}{2022}. \URLprefix \url{https://arxiv.org/abs/2210.07316v3}.
\bibitem[{Mélanie-Becquet et~al.(2024)Mélanie-Becquet, Barré, Seminck, Plancq, Naguib, Pastor, and Poibeau}]{melanie-becquet_booknlp-fr_2024}
\bibinfo{author}{F.~Mélanie-Becquet}, \bibinfo{author}{J.~Barré}, \bibinfo{author}{O.~Seminck}, \bibinfo{author}{C.~Plancq}, \bibinfo{author}{M.~Naguib}, \bibinfo{author}{M.~Pastor}, \bibinfo{author}{T.~Poibeau},
\newblock \bibinfo{title}{{BookNLP}-fr, the {French} {Versant} of {BookNLP}. {A} {Tailored} {Pipeline} for 19th and 20th {Century} {French} {Literature}},
\newblock \bibinfo{journal}{Journal of computational literary studies}  (\bibinfo{year}{2024}). \URLprefix \url{https://tuprints.ulb.tu-darmstadt.de/id/eprint/27396}. \DOIprefix\doi{10.26083/TUPRINTS-00027396}.
\bibitem[{Wang et~al.(2019)Wang, Wang, Chen, Wang, and Kuo}]{SIP-124}
\bibinfo{author}{B.~Wang}, \bibinfo{author}{A.~Wang}, \bibinfo{author}{F.~Chen}, \bibinfo{author}{Y.~Wang}, \bibinfo{author}{C.-C.~J. Kuo},
\newblock \bibinfo{title}{Evaluating word embedding models: methods and experimental results},
\newblock \bibinfo{journal}{APSIPA Transactions on Signal and Information Processing} \bibinfo{volume}{8} (\bibinfo{year}{2019}) \bibinfo{pages}{--}. \DOIprefix\doi{10.1017/ATSIP.2019.12}.
\bibitem[{Timkey and van Schijndel(2021)}]{timkey-van-schijndel-2021-bark}
\bibinfo{author}{W.~Timkey}, \bibinfo{author}{M.~van Schijndel},
\newblock \bibinfo{title}{All bark and no bite: Rogue dimensions in transformer language models obscure representational quality},
\newblock in: \bibinfo{editor}{M.-F. Moens}, \bibinfo{editor}{X.~Huang}, \bibinfo{editor}{L.~Specia}, \bibinfo{editor}{S.~W.-t. Yih} (Eds.), \bibinfo{booktitle}{Proceedings of the 2021 Conference on Empirical Methods in Natural Language Processing}, \bibinfo{publisher}{Association for Computational Linguistics}, \bibinfo{address}{Online and Punta Cana, Dominican Republic}, \bibinfo{year}{2021}, pp. \bibinfo{pages}{4527--4546}. \DOIprefix\doi{10.18653/v1/2021.emnlp-main.372}.
\bibitem[{Mu and Viswanath(2018)}]{mu2018allbutthetop}
\bibinfo{author}{J.~Mu}, \bibinfo{author}{P.~Viswanath}, \bibinfo{title}{All-but-the-top: Simple and effective postprocessing for word representations}, \bibinfo{year}{2018}. \URLprefix \url{https://openreview.net/forum?id=HkuGJ3kCb}.
\bibitem[{Martin et~al.(2020)Martin, Muller, Su{\'a}rez, Dupont, Romary, de~la Clergerie, Seddah, and Sagot}]{martin2020camembert}
\bibinfo{author}{L.~Martin}, \bibinfo{author}{B.~Muller}, \bibinfo{author}{P.~J.~O. Su{\'a}rez}, \bibinfo{author}{Y.~Dupont}, \bibinfo{author}{L.~Romary}, \bibinfo{author}{{\'E}.~V. de~la Clergerie}, \bibinfo{author}{D.~Seddah}, \bibinfo{author}{B.~Sagot},
\newblock \bibinfo{title}{Camembert: a tasty french language mode},
\newblock \bibinfo{journal}{Proceedings of the 58th Annual Meeting of the Association for Computational Linguistics}  (\bibinfo{year}{2020}).
\bibitem[{Underwood et~al.(2022)Underwood, Kiley, Shang, and Vaisey}]{underwood_cohort_2022}
\bibinfo{author}{T.~Underwood}, \bibinfo{author}{K.~Kiley}, \bibinfo{author}{W.~Shang}, \bibinfo{author}{S.~Vaisey},
\newblock \bibinfo{title}{Cohort succession explains most change in literary culture},
\newblock \bibinfo{journal}{Sociological Science} \bibinfo{volume}{9} (\bibinfo{year}{2022}) \bibinfo{pages}{184--205}. \DOIprefix\doi{10.15195/v9.a8}.
\bibitem[{Moretti(2007)}]{moretti_graphs_2007}
\bibinfo{author}{F.~Moretti}, \bibinfo{title}{Graphs, maps, trees: abstract models for literary history}, \bibinfo{publisher}{Verso}, \bibinfo{year}{2007}.
\bibitem[{Compagnon(2013)}]{compagnon_proust_2013}
\bibinfo{author}{A.~Compagnon}, \bibinfo{title}{Proust entre deux siècles}, \bibinfo{publisher}{Éd. du Seuil}, \bibinfo{address}{Paris}, \bibinfo{year}{2013}.
\bibitem[{Lavergne(2009)}]{lavergne_naissance_2009}
\bibinfo{author}{E.~d. Lavergne}, \bibinfo{title}{La naissance du roman policier français: du {Second} {Empire} à la {Première} {Guerre} mondiale}, number~\bibinfo{number}{7} in \bibinfo{series}{Études de littérature des {XXe} et {XXIe} siècles}, \bibinfo{publisher}{Classiques Garnier}, \bibinfo{address}{Paris}, \bibinfo{year}{2009}. \bibinfo{note}{OCLC: ocn436637927}.
\bibitem[{{Jean Barré}(2024)}]{barre_literary_2024}
\bibinfo{author}{{Jean Barré}}, \bibinfo{title}{fr\_literary\_bge\_base}, \bibinfo{year}{2024}. \DOIprefix\doi{10.57967/HF/3255}.

\end{thebibliography}

\appendix

\section{Data, code and model availability}

We have made the data and code available on GitHub\footnote{\url{https://github.com/crazyjeannot/CHR_latent_structures}} and released the fine-tuned model and corpus on HuggingFace.\footnote{\url{https://huggingface.co/crazyjeannot/fr_literary_bge_base}\cite{barre_literary_2024}}

\section{Finetuning supplement}

\begin{figure}[!ht]
    \centering
    \includegraphics[width=10cm]{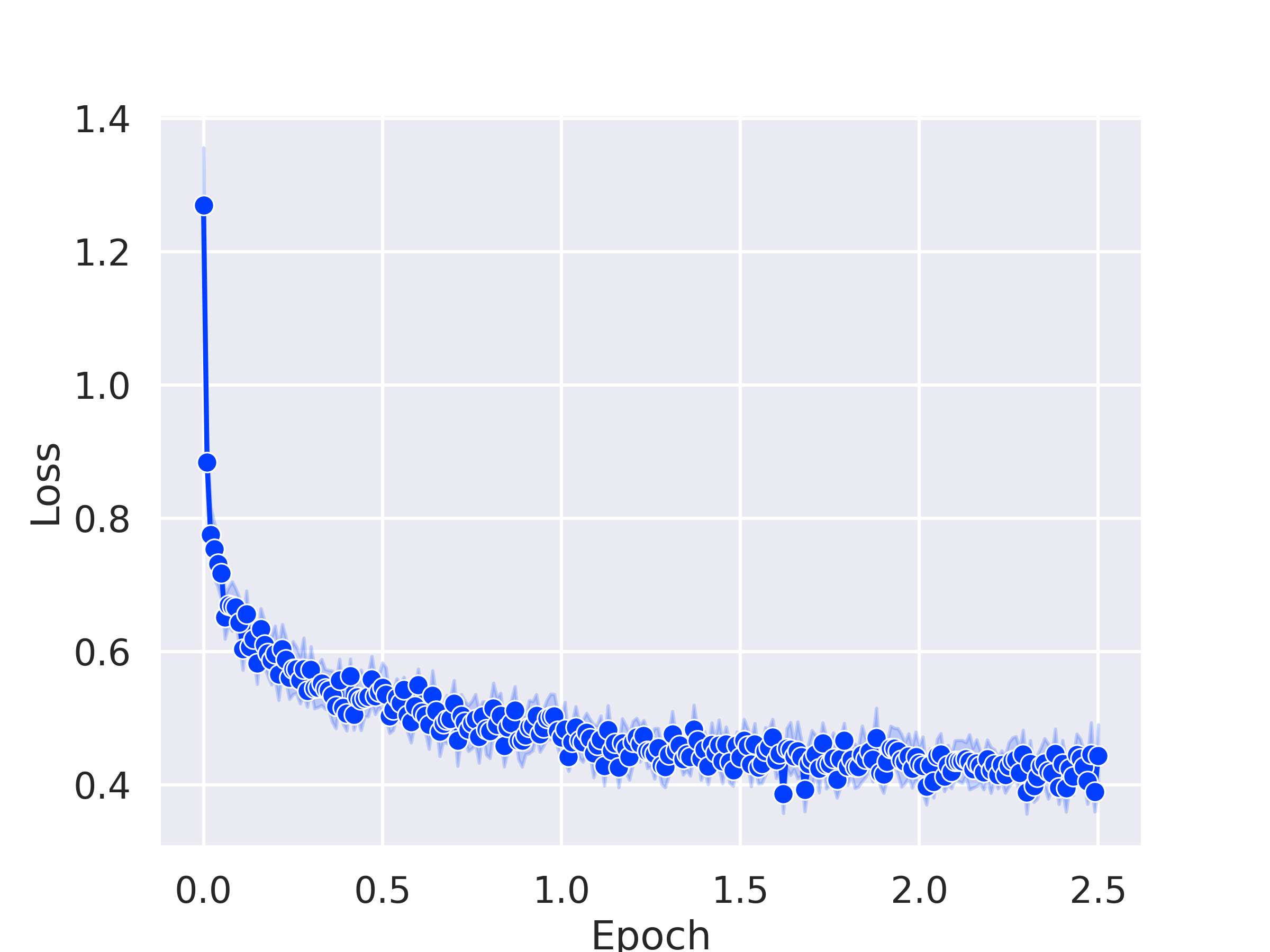}
    \caption{Loss fluctuations during finetuning}
    \label{fig:ft}
\end{figure}

\autoref{fig:ft} shows a rapid decrease in loss at the beginning of the training, from 1.3 to around 0.5 during the first epoch, indicating that the model is quickly learning to distinguish positive passages (subsequent passage) from negative ones (random passage). After this initial drop, the loss decreases more slowly and stabilizes around 0.4 after 2 epochs, suggesting the model is converging. The small fluctuations are likely due to the stochastic nature of the gradient updates, as mini-batches vary in content.

\end{document}